\pdfoutput=1

\documentclass[11pt]{article}

\usepackage[dvipsnames]{xcolor} 
\usepackage[final]{acl}

\usepackage{times}
\usepackage{latexsym}
\usepackage[T1]{fontenc}
\usepackage[utf8]{inputenc}
\usepackage{microtype}
\usepackage{inconsolata}
\usepackage{graphicx}
\usepackage{amsmath}
\usepackage{amssymb}
\usepackage{multirow}
\usepackage{catchfile}
\usepackage{varwidth}
\usepackage{cleveref}
\usepackage[normalem]{ulem}
\usepackage{lipsum}
\usepackage{pifont}
\usepackage[font=small,labelfont=bf]{caption}
\usepackage{float}
\usepackage{booktabs}
\usepackage{subcaption}
\usepackage{natbib}
\usepackage{doi}
\usepackage{appendix}
\usepackage{bm}
\usepackage{wrapfig}
\usepackage{mathtools}

\DeclareMathOperator*{\argmax}{argmax}

\usepackage{amsthm}
\newtheorem{manualtheoreminner}{Theorem}
\newenvironment{manualtheorem}[1]{
  \IfBlankTF{#1}
    {\renewcommand{\themanualtheoreminner}{\unskip}}
    {\renewcommand\themanualtheoreminner{#1}}
  \manualtheoreminner
}{\endmanualtheoreminner}

\usepackage{tikz}
\usetikzlibrary{positioning}
\pgfdeclarelayer{bg}
\pgfsetlayers{bg,main}

\title{On the Mathematical Relationship Between \\ Layer Normalization and Dynamic Activation Functions}

\author{
Felix Stollenwerk \\ AI Sweden \\ \texttt{felix.stollenwerk@ai.se}
}

\begin{document}
\maketitle
\begin{abstract}
Layer normalization (LN) is an essential component of modern neural networks. While many alternative techniques have been proposed, none of them have succeeded in replacing LN so far. The latest suggestion in this line of research is a dynamic activation function called Dynamic Tanh (DyT). Although it is empirically well-motivated and appealing from a practical point of view, it lacks a theoretical foundation. In this work, we shed light on the mathematical relationship between LN and dynamic activation functions. In particular, we derive DyT from the LN variant RMSNorm, and show that a well-defined decoupling in derivative space as well as an approximation are needed to do so. By applying the same decoupling procedure directly in function space, we are able to omit the approximation and obtain the exact element-wise counterpart of RMSNorm, which we call Dynamic Inverse Square Root Unit (DyISRU). We demonstrate numerically that DyISRU reproduces the normalization effect on outliers more accurately than DyT does.
\end{abstract}

\section{Introduction}

Most modern neural network architectures contain normalization layers. These have been shown to have beneficial effects on model training, such as faster, more stable convergence and better results (see e.g. \citet{10056354}). The most widely used normalization layers nowadays, especially in transformers, are layer normalization \citep{ba2016layernormalization} and RMSNorm \citep{NEURIPS2019_1e8a1942}. Both employ activation statistics across the channels (or model dimension) of the neural network. More concretely, consider a hidden representation $x \in \mathbb{R}^C$, corresponding for instance to a single token in the case of an LLM. Layer normalization (LN) acts on $x$ by centering and scaling,
\begin{align}
    y &= \frac{x - \mu}{\sqrt{\sigma^2}}
    \label{eq:layer_normalization}
\end{align}
using the (sample) mean and variance,
\begin{align}
    \mu = \frac{1}{C} \sum_{k=1}^C x_k 
    \ , \quad \
    \sigma^2 = \frac{1}{C} \sum_{k=1}^C \left( x_k - \mu \right)^2
    \label{eq:musigmasq}
\end{align}
Note that $\mu, \sigma \in \mathbb{R}$ and $y \in \mathbb{R}^C$.
RMSNorm is very similar to LN but omits the centering,
\begin{align}
    y &= \frac{x}{\sqrt{\widetilde \sigma^2}}
    \label{eq:rmsnorm}
\end{align}
where
\begin{align}
    \widetilde \sigma^2 
    &= \frac{1}{C} \sum_{k=1}^C x_k^2
    = \frac{ \| x \|^2}{C}
    \label{eq:sigmasq_rmsnorm}
\end{align}
is the uncentered variance and $\| . \|$ denotes the 2-norm.
The impact of normalization on the training of neural networks as well as various alternative methods have been studied in many publications. A concise overview of the most important related work is given in App.~\ref{sec:related work}.
Recently, \citet{Zhu2025DyT} suggested an element-wise, non-linear transformation called Dynamic Tanh (DyT) as a drop-in replacement for LN: 
\begin{align}
    y &= \tanh \left( \alpha x \right)
    \label{eq:introduction_dyt}
\end{align} 
It uses a learnable parameter $\alpha \in \mathbb{R}$, instead of relying on activation statistics like the traditional normalization methods discussed above.
DyT is empirically well-motivated, as it was shown to resemble LN in the sense that it linearly transforms small values of $x$ while squashing large values.  
However, the authors did not provide a theoretical justification for the observed similarity. 
The aim of our work is to fill this gap and to enhance the theoretical understanding of the relationship between LN and dynamic activation functions. 

The paper is structured as follows. 
In Sec.~\ref{sec:dyt}, we show that DyT can be mathematically derived from RMSNorm using a well-defined decoupling in derivative space.
Sec.~\ref{sec:dyisru} discusses a similar element-wise transformation called \textit{Dynamic Inverse Square Root Unit (DyISRU)}, which emerges from a decoupling procedure directly in function space.
Sec.~\ref{sec:simulations} demonstrates that DyISRU is more similiar to normalization than DyT with regard to its effect on outliers.
Our conclusions are presented in Sec.~\ref{sec:conclusions}.

\section{Dynamic Tanh (DyT)}
\label{sec:dyt}

In this section, we provide a mathematical derivation of the DyT function as an element-wise approximation of LN.
Note that our starting point is actually RMSNorm, reflecting the fact that dynamic activation functions like Eq.~(\ref{eq:introduction_dyt}) do not involve any subtractions for the purpose of centering.
The derivation is done in three steps: 
(i) The derivative of the RMSNorm output with respect to its input is computed, resulting in a system of differential equations.
(ii) The system of differential equations is decoupled and thereby reduced to a single equation.
(iii) The decoupled differential equation is solved, leading to the DyT function.
The process is illustrated in Fig.~\ref{fig:tikz}.
\begin{figure*}[ht]
\centering
\begin{tikzpicture}[
blacknode/.style={rectangle, draw=black!60, fill=black!5, very thick, minimum size=10mm},
greennode/.style={rectangle, draw=green!60, fill=green!5, very thick, minimum size=10mm},
bluenode/.style={rectangle, draw=blue!60, fill=blue!5, very thick, minimum size=10mm},
rednode/.style={rectangle, draw=red!60, fill=red!5, very thick, minimum size=10mm},
nonode/.style={rectangle, draw=white!60, fill=white!5, very thick, minimum size=10mm},
]

\node[blacknode, label={[xshift=-20mm, yshift=-8mm]\textbf{RMSNorm}}] (node1) at (0,0) {$y_i = \frac{x_i}{\sqrt{\widetilde \sigma^2}}$};
\node[nonode]    (node2) at (0,-2) {};
\node[rednode, label={[xshift=4mm, yshift=1mm]\textbf{T1}}]   (node3) at (0,-4.5) {$\frac{\partial y_i}{\partial x_j} = \frac{\sqrt{C}}{\| x \|} \left( \delta_{ij} - \frac{y_i y_j}{C} \right)$};
\node[rednode, label={[xshift=-4mm, yshift=13mm]\textbf{T2}}]   (node4) at (9.2,-4.5) {$\frac{d y_i}{d x_i} = \sqrt{\alpha C} \left( 1 - \frac{y_i^2}{C} \right)$};
\node[rednode, label={[xshift=24mm, yshift=-8mm]\textbf{\textcolor{red}{DyT}}}]   (node5) at (9.2,-1.5) {$y_i = \sqrt{C} \tanh \left( \alpha x_i \right)$};
\node[bluenode, label={[xshift=-22mm, yshift=-12mm]\textbf{T3}}] (node11) at (4.75,0) {$y_i = \sqrt{C} \frac{x_i}{\sqrt{ \| x_{\backslash i} \|^2 + x_i^2}}$};
\node[bluenode, label={[xshift=24mm, yshift=-8mm]\textbf{\textcolor{blue}{DyISRU}}}] (node12) at (9.5,0) {$y_i = \sqrt{C} \frac{x_i}{\sqrt{ \beta + x_i^2}}$};

\draw[->, line width = 0.3mm] (node1.south) -- (node3.north) node[pos=0.72,left,fill=white] {$\frac{\partial}{\partial x_j}$};
\draw[->, line width = 0.3mm] (node3.east)  -- (node4.west) node[midway,above] {\small decouple};
\draw[->, line width = 0.3mm] (node4.north) -- (node5.south) node[midway,right,fill=white] {$\int dx_i$};
\draw[->, line width = 0.3mm] (node1.east)  -- (node11.west)node[midway,above] {\small rearrange};
\draw[->, line width = 0.3mm] (node11.east) -- (node12.west) node[midway,above] {\small decouple};
\begin{pgfonlayer}{bg}
\draw[dashed] (-2.8,-3) -- (12.8,-3);
\end{pgfonlayer}
\end{tikzpicture}
\caption{Illustration of how to obtain the dynamic activation functions DyT (red) and DyISRU (blue) from RMSNorm (black). The labels T1, T2, T3 indicate the application of our theorems. The dashed line differentiates between function space ($y_i$) above and derivative space ($\frac{\partial y_i}{\partial x_j}$) below.}
\label{fig:tikz}
\end{figure*}

\begin{manualtheorem}{1}[RMSNorm Derivative]
Let $x \in \mathbb{R}^C$ and 
\begin{align}
    y &= \frac{x}{\sqrt{\widetilde \sigma^2}}
    \tag{\ref{eq:rmsnorm}}
\end{align}
with the uncentered variance of $x$,
\begin{align}
    \widetilde \sigma^2 &= \frac{1}{C} \sum_{k=1}^C x_k^2 = \frac{\| x \|^2}{C} 
    \tag{\ref{eq:sigmasq_rmsnorm}}
\end{align}
Then $\forall ~i,j \in [1, \ldots, C]$:
\begin{align}
    \frac{\partial y_i}{\partial x_j} 
    &= \frac{\sqrt{C}}{\| x \|} \left( \delta_{ij} - \frac{y_i y_j}{C} \right)
    \label{eq:theorem1_rms}
\end{align}
\end{manualtheorem}
While this result has been obtained before \cite{pmlr-v119-xiong20b, takase2025spikemorestabilizingpretraining}, we provide a proof using our notation in App.~\ref{app:theorem1}.

\paragraph{Decoupling}
The partial derivatives of $y_i$ in Eq.~(\ref{eq:theorem1_rms}) depend on the different components of $x$ in two ways. Firstly, through the off-diagonal elements ($i \neq j$) of the Jacobian. Secondly, all components $x_k$ enter the expression for $\| x \|$.
We can remove these cross-dependencies and decouple the system of differential equations by ignoring the off-diagonal terms and replacing the norm of $x$ by a constant:
\begin{align}
    y_i y_j &\to 0 \quad \forall \ i \neq j \label{eq:dyt_decoupling_1} \\
    \| x \| &\to \frac{1}{\sqrt{\alpha}}
    \label{eq:dyt_assumption}
\end{align}
With these replacements, Eq.~(\ref{eq:theorem1_rms}) becomes
\begin{align}
    \frac{\partial y_i}{\partial x_j} 
    &= \begin{cases}
        \sqrt{\alpha C} \left( 1 - \frac{y_i^2}{C} \right) & \text{ if } i = j \\
        0 & \text{ if } i \neq j
    \end{cases}
    \label{eq:theorem1_rms_decoupled_cases}
\end{align}
We write the single differential equation on the diagonal succinctly as
\begin{align}
    \frac{dy_i}{dx_i} 
    &= \sqrt{\alpha C} \left( 1 - \frac{y_i^2}{C} \right)
    \label{eq:theorem1_rms_decoupled}
\end{align}
The following theorem shows that the DyT function can be obtained by solving Eq.~(\ref{eq:theorem1_rms_decoupled}).
\begin{manualtheorem}{2}[DyT]
The differential equation 
\begin{align}
    \frac{dy_i}{dx_i} 
    &= \sqrt{\alpha C} \left( 1 - \frac{y_i^2}{C} \right) \tag{\ref{eq:theorem1_rms_decoupled}}
\end{align}
together with the initial condition
\begin{align}
    y_i(x_i = 0) &= 0 \label{eq:bc2}
\end{align}
is solved by the function
\begin{align}
    y_i &= \sqrt{C} \cdot \tanh \left( \alpha x_i \right)
    \label{eq:dyt}
\end{align}
\end{manualtheorem}
The proof can be found in App.~\ref{app:theorem2}.
Eq.~(\ref{eq:dyt}) represents the (scaled) DyT function.  
In contrast to the original formulation of DyT, Eq.~(\ref{eq:introduction_dyt}), it explicitly contains the minimum and maximum value of $y_i$ in terms of the scaling factor $\sqrt{C}$.

\section{Dynamic Inverse Square Root Unit (DyISRU)}
\label{sec:dyisru}

The results of the previous section raise the question whether it is possible to find an element-wise transformation akin to DyT that avoids both the detour through derivative space and neglecting off-diagonal elements of the Jacobian, Eq.~(\ref{eq:dyt_decoupling_1}). Such a function could be a more direct and accurate element-wise replacement for RMSNorm.
To answer this question, we first rearrange the RMSNorm transformation, Eq.~(\ref{eq:rmsnorm}), and then decouple the components as before, see Fig.~\ref{fig:tikz}.
\begin{manualtheorem}{3}[RMSNorm Rearrangement]
Let $x \in \mathbb{R}^C$ and 
\begin{align}
y_i &:= \frac{x_i}{\sqrt{\widetilde \sigma^2}}
\tag{\ref{eq:rmsnorm}}
\end{align}
be the output of the RMSNorm with
\begin{align}
    \widetilde \sigma^2 
    &= \frac{1}{C} \sum_{k=1}^C x_k^2
    = \frac{1}{C} \| x \|^2
    \tag{\ref{eq:sigmasq_rmsnorm}}
\end{align}
being the uncentered variance of $x$.
Eq.~(\ref{eq:rmsnorm}) can be written as
\begin{align}
y_i 
 &= \sqrt{C} \cdot \frac{x_i}{\sqrt{\| x_{\backslash i} \|^2 + x_i^2}} \label{eq:rmsnorm_alternative}
\end{align}
where
\begin{align}
    \| x_{\backslash i} \| := \sqrt{ \sum_{k \neq i} x_k^2 }
    \label{eq:definition_x_not_i}
\end{align}
is the 2-norm of the vector $x_{\backslash i} \in \mathbb{R}^{C-1}$ comprised of all channels but $i$.
\end{manualtheorem}
The proof can be found in App.~\ref{app:theorem3}.

\paragraph{Decoupling}

The rearranged formulation of RMSNorm in Eq.~(\ref{eq:rmsnorm_alternative}) is convenient as the dependency on the components $x_j$ with $j \neq i$ is isolated by means of $\| x_{\backslash i} \|$. An element-wise transformation can thus easily be obtained using the replacement
\begin{align}
    \| x_{\backslash i} \| &\to \beta
    \label{eq:dyisru_assumption}
\end{align}
where $\beta$ is a global constant. It reads
\begin{align}
    y_i
    &= \sqrt{C} \cdot \frac{x_i}{\sqrt{ \beta + x_i^2}}
    \label{eq:dyisru}
\end{align}
We call this function Dynamic Inverse Square Root Unit (DyISRU), as it is proportional to the ISRU activation function \citep{carlile2017improvingdeeplearninginverse}, see App.~\ref{app:isru} for details. Like the (scaled) DyT function in Eq.~(\ref{eq:dyt}), it explicitly contains the minimum and maximum values of $y_i$ in terms of $\sqrt{C}$. 
In App.~\ref{app:practical_considerations}, we provide some considerations on how to use it in practice.

\paragraph{Comparison}

\begin{figure}[hb!]
\centering
\includegraphics[scale=0.5]{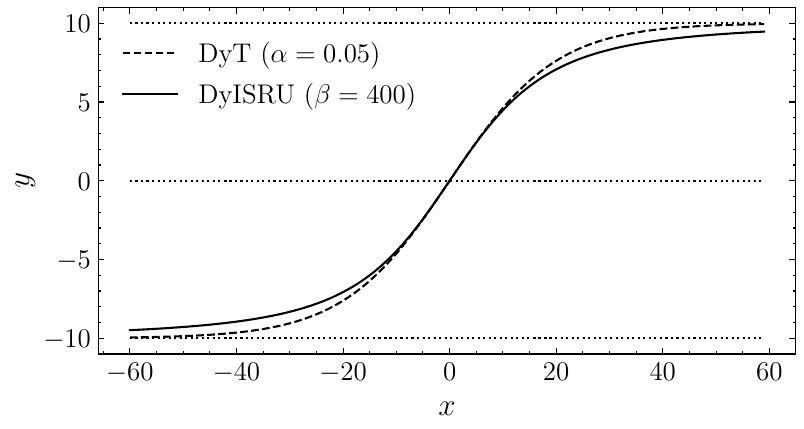}
\caption{Functions DyT from Eq.~(\ref{eq:dyt}) and DyISRU from Eq.~(\ref{eq:dyisru}) with parameters $\alpha = 0.05$ and $\beta = 400$ such that the derivatives at $x=0$, namely $\alpha$ and $1/\sqrt{\beta}$, match. The dotted lines correspond to the extrema $y = \pm \sqrt{C}$.}
\label{fig:solutions}
\end{figure}

The two discussed element-wise transformations, DyT from Eq.~(\ref{eq:dyt}) and DyISRU from Eq.~(\ref{eq:dyisru}), are compared in Fig.~\ref{fig:solutions}.
Their shapes are quite similar, but DyT converges faster to the extrema ($\pm \sqrt{C}$) than DyISRU.
Both dynamic activation functions employ learnable parameters instead of relying on activation statistics. 
However, DyT's $\alpha$ globally models the inverse uncentered variance of $x$ including all channels, Eq.~(\ref{eq:dyt_assumption}), 
while DyISRU's $\beta$ globally models the uncentered variance of $x$ including all channels but the transformed one, Eq.~(\ref{eq:dyisru_assumption}).
More importantly, unlike DyISRU, DyT can be considered an approximation as it ignores the off-diagonal entries of the Jacobian, Eq.~(\ref{eq:dyt_decoupling_1}).
This suggests that DyISRU more closely resembles normalization in comparison with DyT. In the next section, we will see that this is indeed the case.

\section{Outlier Simulation}
\label{sec:simulations}

In order to compare RMSNorm and the considered dynamic activation functions, we first simulate data with outliers of different degrees and apply normalization. Afterwards, we employ DyT and DyISRU with optimal parameters $\alpha$ and $\beta$ to investigate how well they describe the data. 

\subsection{Data: RMSNorm}

We assume $C$ channels and take a normally distributed sample of $C$ values $x = (x_1, x_2, \ldots, x_C)$ with zero mean and standard deviation $\sigma$:
\begin{align}
    x \sim \mathcal{N}(0, \sigma^2)
    \label{eq:outlier_simulation_normal_distribution}
\end{align}
The parameters are set to arbitrary values, $C=100$ and $\sigma = 2$.
First, we apply RMSNorm and compute $y$ according to Eq.~(\ref{eq:rmsnorm}). 
Next, we simulate outliers by iteratively increasing the largest value of $x$ in steps of $5$:
\begin{equation}
    x_o \to x_o + 5 \cdot S 
    \quad \text{with} \quad 
    o = \argmax_{k} x_k
    \label{eq:outliers}
\end{equation}
where $S$ denotes the number of steps. At each step, RMSNorm is applied to the vector $x$ to yield the output $y$. 
Repeating the process up to $S=9$ yields the top plot of Fig.~\ref{fig:simulation_fits}. 
In accordance with \citet{Zhu2025DyT}, we observe that (i) the slope of the linear function $y_i(x_i)$ decreases with the variance of $x$ and (ii) the outliers follow a non-linear function. The larger the outlier $x_o$, the more squashed is the function.

\subsection{Fit: DyT and DyISRU}

\begin{figure}[t]
\centering
\includegraphics[scale=0.5]{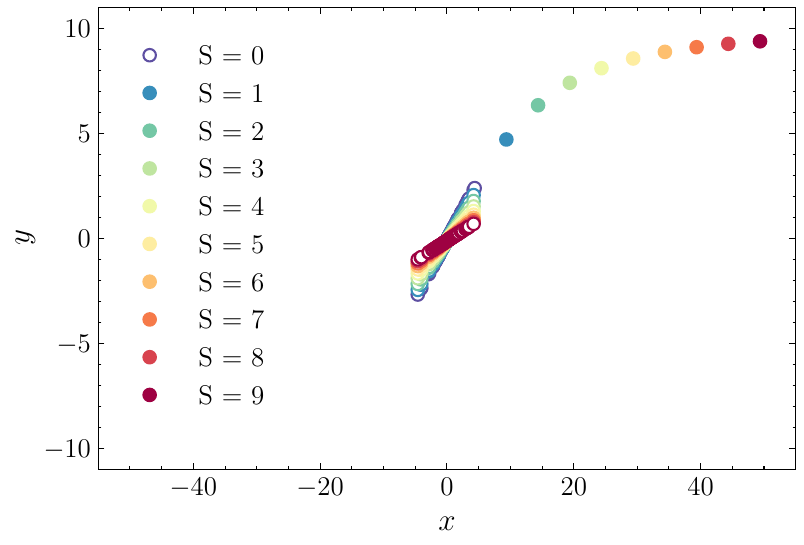} \\
\includegraphics[scale=0.5]{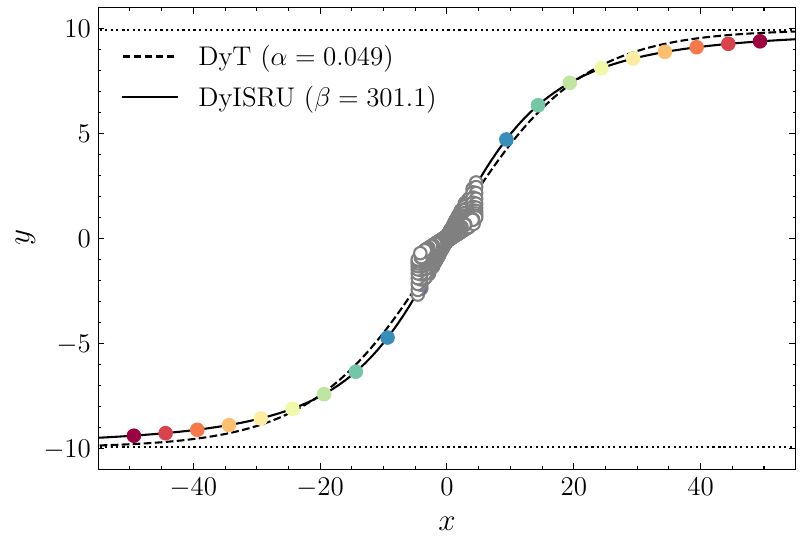} \\
\includegraphics[scale=0.5]{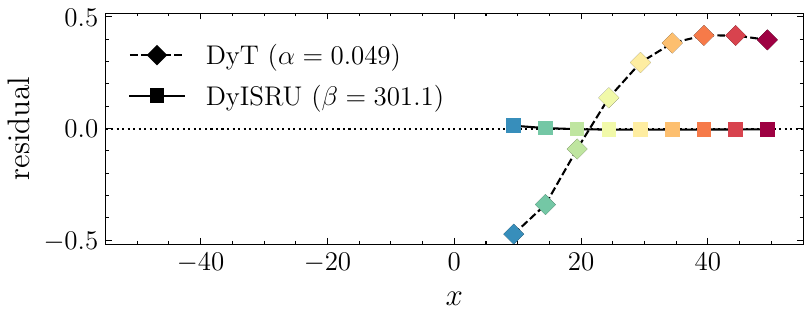}
\caption{\textit{Top:} Stepwise outlier simulation. The sample $x$ and is plotted against its normalized counterpart $y$, with outliers of different degrees (filled circles) as defined by Eq.~(\ref{eq:outliers}). \textit{Center:} Functions DyT and DyISRU with optimal parameters $\alpha$ and $\beta$, respectively, fitted on the outliers shown as colored, filled circles. The non-outlier data are shown as gray, empty circles. \textit{Bottom:} Residuals of the functions DyT and DyISRU with respect to the outlier data. As the residuals are antisymmetric (like the data and the functions), only positive outliers are displayed for the sake of simplicity.}
\label{fig:simulation_fits}
\end{figure}

In the next step, we aim to describe the simulated data by DyT and DyISRU as defined in Eq.~(\ref{eq:dyt}) and (\ref{eq:dyisru}), respectively. We only use the outliers (filled circles in Fig.~\ref{fig:simulation_fits}) as data points for the fit\footnote{In practice, we use the mirrored data points $(-x, -y)$ as well for the sake of numerical stability.}, since it is primarily those we want to reproduce the normalization behavior for. 
Performing fits yields the optimal parameters
\begin{align}
    \alpha = 0.049 
    \quad \text{and} \quad
    \beta = 301.1 
\end{align}
for DyT and DyISRU, respectively. Both fitted functions are displayed in Fig.~\ref{fig:simulation_fits} together with the data and the residuals.
We find that DyISRU describes the normalization data more accurately than DyT does. The mean absolute residuals are $0.33$ for DyT and $<0.01$ for DyISRU. This is consistent with our theoretical finding that DyT and DyISRU correspond to approximate and exact element-wise counterparts of RMSNorm, respectively.
Finally, we note that our results generalize to other distributions than the one used above, Eq.~(\ref{eq:outlier_simulation_normal_distribution}). Further details are provided in App.~\ref{app:outlier_simulation_generalization}.

\section{Conclusions}
\label{sec:conclusions}

This work provides a theoretical foundation for the empirically observed similarity of DyT and LN. We have detailed how dynamic activation functions emerge mathematically from RMSNorm by promoting channel-specific terms to a global, learnable parameter designed to describe outliers (decoupling).
In addition, our analysis reveals that the exact element-wise counterpart to RMSNorm is given by a transformation called Dynamic Inverse Square Root Unit (DyISRU), while DyT can be considered an approximation. 
The code used to reproduce our results is available at \url{https://github.com/flxst/dynamic-activation-functions}.

\section{Limitations}

While our work advances the theoretical understanding of dynamic activation functions, we did not conduct any experiments. Hence, we cannot make any statements about the practical implications of our work. In particular, it remains to be seen whether there is a notable difference between DyT and DyISRU in terms of model performance and with regard to the sensitivity on the learnable parameters' initial values, which has been reported especially for LLMs \cite{Zhu2025DyT}.

\bibliography{custom}

\begin{thebibliography}{16}
\expandafter\ifx\csname natexlab\endcsname\relax\def\natexlab#1{#1}\fi

\bibitem[{Ba et~al.(2016)Ba, Kiros, and Hinton}]{ba2016layernormalization}
Jimmy~Lei Ba, Jamie~Ryan Kiros, and Geoffrey~E. Hinton. 2016.
\newblock \href {http://arxiv.org/abs/1607.06450} {Layer normalization}.

\bibitem[{Brock et~al.(2021{\natexlab{a}})Brock, De, and Smith}]{brock2021characterizingsignalpropagationclose}
Andrew Brock, Soham De, and Samuel~L. Smith. 2021{\natexlab{a}}.
\newblock \href {http://arxiv.org/abs/2101.08692} {Characterizing signal propagation to close the performance gap in unnormalized resnets}.

\bibitem[{Brock et~al.(2021{\natexlab{b}})Brock, De, Smith, and Simonyan}]{pmlr-v139-brock21a}
Andy Brock, Soham De, Samuel~L Smith, and Karen Simonyan. 2021{\natexlab{b}}.
\newblock \href {https://proceedings.mlr.press/v139/brock21a.html} {High-performance large-scale image recognition without normalization}.
\newblock In \emph{Proceedings of the 38th International Conference on Machine Learning}, volume 139 of \emph{Proceedings of Machine Learning Research}, pages 1059--1071. PMLR.

\bibitem[{Carlile et~al.(2017)Carlile, Delamarter, Kinney, Marti, and Whitney}]{carlile2017improvingdeeplearninginverse}
Brad Carlile, Guy Delamarter, Paul Kinney, Akiko Marti, and Brian Whitney. 2017.
\newblock \href {http://arxiv.org/abs/1710.09967} {Improving deep learning by inverse square root linear units (isrlus)}.

\bibitem[{De and Smith(2020)}]{NEURIPS2020_e6b738ec}
Soham De and Sam Smith. 2020.
\newblock \href {https://proceedings.neurips.cc/paper_files/paper/2020/file/e6b738eca0e6792ba8a9cbcba6c1881d-Paper.pdf} {Batch normalization biases residual blocks towards the identity function in deep networks}.
\newblock In \emph{Advances in Neural Information Processing Systems}, volume~33, pages 19964--19975. Curran Associates, Inc.

\bibitem[{Dugas et~al.(2000)Dugas, Bengio, B{\'e}lisle, Nadeau, and Garcia}]{dugas2000incorporating}
Charles Dugas, Yoshua Bengio, Fran{\c{c}}ois B{\'e}lisle, Claude Nadeau, and Ren{\'e} Garcia. 2000.
\newblock Incorporating second-order functional knowledge for better option pricing.
\newblock \emph{Advances in neural information processing systems}, 13.

\bibitem[{He and Hofmann(2024)}]{he2024simplifyingtransformerblocks}
Bobby He and Thomas Hofmann. 2024.
\newblock \href {http://arxiv.org/abs/2311.01906} {Simplifying transformer blocks}.

\bibitem[{Huang et~al.(2023)Huang, Qin, Zhou, Zhu, Liu, and Shao}]{10056354}
Lei Huang, Jie Qin, Yi~Zhou, Fan Zhu, Li~Liu, and Ling Shao. 2023.
\newblock \href {https://doi.org/10.1109/TPAMI.2023.3250241} {Normalization techniques in training dnns: Methodology, analysis and application}.
\newblock \emph{IEEE Transactions on Pattern Analysis and Machine Intelligence}, 45(8):10173--10196.

\bibitem[{Klambauer et~al.(2017)Klambauer, Unterthiner, Mayr, and Hochreiter}]{NIPS2017_5d44ee6f}
G\"{u}nter Klambauer, Thomas Unterthiner, Andreas Mayr, and Sepp Hochreiter. 2017.
\newblock \href {https://proceedings.neurips.cc/paper_files/paper/2017/file/5d44ee6f2c3f71b73125876103c8f6c4-Paper.pdf} {Self-normalizing neural networks}.
\newblock In \emph{Advances in Neural Information Processing Systems}, volume~30. Curran Associates, Inc.

\bibitem[{Lyu et~al.(2023)Lyu, Li, and Arora}]{lyu2023understandinggeneralizationbenefitnormalization}
Kaifeng Lyu, Zhiyuan Li, and Sanjeev Arora. 2023.
\newblock \href {http://arxiv.org/abs/2206.07085} {Understanding the generalization benefit of normalization layers: Sharpness reduction}.

\bibitem[{Ni et~al.(2024)Ni, Guo, Jia, and Huang}]{pmlr-v235-ni24b}
Yunhao Ni, Yuxin Guo, Junlong Jia, and Lei Huang. 2024.
\newblock \href {https://proceedings.mlr.press/v235/ni24b.html} {On the nonlinearity of layer normalization}.
\newblock In \emph{Proceedings of the 41st International Conference on Machine Learning}, volume 235 of \emph{Proceedings of Machine Learning Research}, pages 37957--37998. PMLR.

\bibitem[{Takase et~al.(2025)Takase, Kiyono, Kobayashi, and Suzuki}]{takase2025spikemorestabilizingpretraining}
Sho Takase, Shun Kiyono, Sosuke Kobayashi, and Jun Suzuki. 2025.
\newblock \href {http://arxiv.org/abs/2312.16903} {Spike no more: Stabilizing the pre-training of large language models}.

\bibitem[{Xiong et~al.(2020)Xiong, Yang, He, Zheng, Zheng, Xing, Zhang, Lan, Wang, and Liu}]{pmlr-v119-xiong20b}
Ruibin Xiong, Yunchang Yang, Di~He, Kai Zheng, Shuxin Zheng, Chen Xing, Huishuai Zhang, Yanyan Lan, Liwei Wang, and Tieyan Liu. 2020.
\newblock \href {https://proceedings.mlr.press/v119/xiong20b.html} {On layer normalization in the transformer architecture}.
\newblock In \emph{Proceedings of the 37th International Conference on Machine Learning}, volume 119 of \emph{Proceedings of Machine Learning Research}, pages 10524--10533. PMLR.

\bibitem[{Zhang and Sennrich(2019)}]{NEURIPS2019_1e8a1942}
Biao Zhang and Rico Sennrich. 2019.
\newblock \href {https://proceedings.neurips.cc/paper_files/paper/2019/file/1e8a19426224ca89e83cef47f1e7f53b-Paper.pdf} {Root mean square layer normalization}.
\newblock In \emph{Advances in Neural Information Processing Systems}, volume~32. Curran Associates, Inc.

\bibitem[{Zhang et~al.(2019)Zhang, Dauphin, and Ma}]{zhang2019fixupinitializationresiduallearning}
Hongyi Zhang, Yann~N. Dauphin, and Tengyu Ma. 2019.
\newblock \href {http://arxiv.org/abs/1901.09321} {Fixup initialization: Residual learning without normalization}.

\bibitem[{Zhu et~al.(2025)Zhu, Chen, He, LeCun, and Liu}]{Zhu2025DyT}
Jiachen Zhu, Xinlei Chen, Kaiming He, Yann LeCun, and Zhuang Liu. 2025.
\newblock Transformers without normalization.
\newblock In \emph{Proceedings of the IEEE/CVF Conference on Computer Vision and Pattern Recognition (CVPR)}.

\end{thebibliography}

\appendix
\section{Related Work}
\label{sec:related work}

Our work is closely related to \citet{Zhu2025DyT} and the idea of using dynamic activation functions to replace layer normalization in neural networks. However, other alternatives to normalization have been suggested in the literature. The arguably closest relatives to dynamic activation functions are self-normalizing neural networks \citep{NIPS2017_5d44ee6f}, which use the special SELU activation function to enforce implicit normalization. Other notable work seeks to compensate for the absence of normalization by means of special initialization schemes \citep{zhang2019fixupinitializationresiduallearning, NEURIPS2020_e6b738ec} and additional techniques \citep{brock2021characterizingsignalpropagationclose, pmlr-v139-brock21a}. Furthermore, approaches specific to transformers have been proposed, like alternative transformer block architectures that dispense with normalization \citep{he2024simplifyingtransformerblocks}.
While we provide insights into how layer normalization relates specifically to dynamic activation functions, the impact of layer normalization has been analyzed with regard to many different aspects, including but not limited to non-linearity \citep{pmlr-v235-ni24b} and generalization \citep{lyu2023understandinggeneralizationbenefitnormalization}.

\section{Theorem Proofs}

\subsection{Theorem 1}
\label{app:theorem1}

\begin{manualtheorem}{1}[RMSNorm Derivative]
Let $x \in \mathbb{R}^C$ and 
\begin{align}
    y &= \frac{x}{\sqrt{\widetilde \sigma^2}}
    \tag{\ref{eq:rmsnorm}}
\end{align}
with the uncentered variance of $x$,
\begin{align}
    \widetilde \sigma^2 &= \frac{1}{C} \sum_{k=1}^C x_k^2 = \frac{\| x \|^2}{C} 
    \tag{\ref{eq:sigmasq_rmsnorm}}
\end{align}
Then $\forall ~i,j \in [1, \ldots, C]$:
\begin{align}
    \frac{\partial y_i}{\partial x_j} 
    &= \frac{\sqrt{C}}{\| x \|} \left( \delta_{ij} - \frac{y_i y_j}{C} \right)
    \tag{\ref{eq:theorem1_rms}}
\end{align}
\end{manualtheorem}
\begin{proof}
We start from Eq.~(\ref{eq:rmsnorm}) and compute the derivative of $y_i$ with respect to $x_j$:
\begin{align}
    \frac{\partial y_i}{\partial x_j} 
    &= \frac{\partial}{\partial x_j} \left( \frac{x_i}{\sqrt{\widetilde \sigma^2}} \right) \label{eq:derivative_1_rms}
\end{align}
Defining 
\begin{align}
    f = x_i
    \ \qquad \
    g = \sqrt{\widetilde \sigma^2} 
    \label{eq:fg_rms}
\end{align}
and using the shorthand notation
\begin{align}
    f' := \frac{\partial f}{\partial x_j} 
    \ \qquad \
    g' := \frac{\partial g}{\partial x_j}
\end{align}
the quotient rule states
\begin{align}
    \frac{\partial y_i}{\partial x_j} 
    &= \frac{f' g - f g'}{g^2}
    \label{eq:quotient_rule}
\end{align}
We compute the derivatives in Eq.~(\ref{eq:quotient_rule}):
\begin{align}
    f' &\stackrel{(\ref{eq:fg_rms})}{=} \frac{\partial}{\partial x_j} x_i \nonumber \\
       &= \delta_{ij} \label{eq:fprime_rms}
\end{align}
and
\begin{align}
    g' &\stackrel{(\ref{eq:fg_rms})}{=} \frac{\partial}{\partial x_j} \sqrt{\widetilde \sigma^2} \nonumber \\
       &= \frac{1}{2 \sqrt{\widetilde \sigma^2}} \cdot \frac{\partial}{\partial x_j} \left( \widetilde \sigma^2 \right) \nonumber \\
       &\stackrel{(\ref{eq:sigmasq_rmsnorm})}{=} \frac{1}{2 \sqrt{\widetilde \sigma^2}} \cdot \frac{\partial}{\partial x_j} \left( \frac{1}{C} \sum_{k=1}^C x_k^2 \right) \nonumber \\
       &= \frac{1}{2 C \sqrt{\widetilde \sigma^2}} \cdot \frac{\partial}{\partial x_j} \left( x_j^2 + \sum_{k \neq j}^C x_k^2 \right) \nonumber \\
       &= \frac{1}{C \sqrt{\widetilde \sigma^2}} \cdot x_j \nonumber \\
       &\stackrel{(\ref{eq:rmsnorm})}{=} \frac{y_j}{C} 
       \label{eq:gprime_rms}
\end{align}
Inserting Eqs.~(\ref{eq:fprime_rms}) and (\ref{eq:gprime_rms}) into Eq.~(\ref{eq:quotient_rule}), we get
\begin{align}
    \frac{\partial y_i}{\partial x_j} 
    &\stackrel{(\ref{eq:quotient_rule})}{=} \frac{f' g - f g'}{g^2} \nonumber \\
    &\stackrel{(\ref{eq:fprime_rms}, \ref{eq:gprime_rms})}{=}
    \frac{\delta_{ij} \cdot \sqrt{\widetilde \sigma^2} - x_i \cdot \frac{y_j}{C}}{\widetilde \sigma^2} \nonumber \\
    &\stackrel{(\ref{eq:rmsnorm})}{=} \frac{\delta_{ij} \cdot \sqrt{\widetilde \sigma^2} - \frac{1}{C} \sqrt{\widetilde \sigma^2} \cdot y_i y_j}{\widetilde \sigma^2} \nonumber \\
    &= \frac{1}{\sqrt{\widetilde \sigma^2}} \left( \delta_{ij} - \frac{y_i y_j}{C} \right) \label{eq:yprime_final_rms} \\
    &\stackrel{(\ref{eq:sigmasq_rmsnorm})}{=} \frac{\sqrt{C}}{\| x \|} \left( \delta_{ij} - \frac{y_i y_j}{C} \right)
\end{align}
\end{proof}

\subsection{Theorem 2}
\label{app:theorem2}

\begin{manualtheorem}{2}[DyT]
The differential equation 
\begin{align}
    \frac{dy_i}{dx_i} 
    &= \sqrt{\alpha C} \left( 1 - \frac{y_i^2}{C} \right) \tag{\ref{eq:theorem1_rms_decoupled}}
\end{align}
together with the initial condition
\begin{align}
    y_i(x_i = 0) &= 0 \tag{\ref{eq:bc2}}
\end{align}
is solved by the function
\begin{align}
    y_i &= \sqrt{C} \cdot \tanh \left( \alpha x_i \right)
    \tag{\ref{eq:dyt}}
\end{align}
\end{manualtheorem}

\begin{proof}
For the sake of readability, we temporarily drop the channel index $i$, i.e. we use $x_i \to x$ and $y_i \to y$. An additional slight rearrangement leads to
\begin{align}
    \frac{dy}{dx} 
    &= \frac{\sqrt{\alpha}}{\sqrt{C}} \left( C - y^2 \right) 
\end{align}
We separate the variables:
\begin{align}
    \frac{\sqrt{\alpha}}{\sqrt{C}} \cdot dx
    &= \frac{dy}{C - y^2} \nonumber \\
    &= \frac{dy}{(\sqrt{C} - y) (\sqrt{C} + y)} \nonumber \\
    &= \frac{1}{2 \sqrt{C}} \frac{\sqrt{C} - y + \sqrt{C} + y}{(\sqrt{C} - y) (\sqrt{C} + y)} dy \nonumber \\
    &= \frac{1}{2 \sqrt{C}} \left( \frac{1}{\sqrt{C} + y} + \frac{1}{\sqrt{C} - y} \right) dy
\end{align}
Integration yields
\begin{align}
    \frac{1}{2 \sqrt{C}} \log \left( \frac{\sqrt{C} + y}{\sqrt{C} - y} \right) &= \frac{\alpha x}{\sqrt{C}} + \frac{c}{2 \sqrt{C}} \nonumber \\
    \frac{\sqrt{C} + y}{\sqrt{C} - y} &= \exp \left( 2 \alpha x + c \right) 
\end{align}
where $c$ is an integration constant.
Defining $Q := \exp \left( 2 \alpha x + c \right)$, we get
\begin{align}
    \frac{\sqrt{C} + y}{\sqrt{C} - y} &= Q \nonumber \\
    \sqrt{C} + y &= \left( \sqrt{C} - y \right) Q \nonumber \\
    \sqrt{C} + y &= \sqrt{C} Q - Q y \nonumber \\
    Q y + y &= \sqrt{C} \left( Q - 1 \right) \nonumber \\
    (Q + 1) y &= \sqrt{C} \left( Q - 1 \right) \nonumber \\
    y &= \sqrt{C} \cdot \frac{Q-1}{Q+1}
\end{align}
Replacing $Q$ again, and using $A = \exp (c)$, yields
\begin{align}
    y_i &= \sqrt{C} \cdot \frac{A \exp \left( 2 \alpha x_i \right)-1}{A \exp \left( 2 \alpha x_i \right)+1}
\end{align}
Note that in the last equation, we have reintroduced the channel index $i$.
We enforce the initial condition from Eq.~(\ref{eq:bc2})
which requires $A = 1$. This leads to the scaled DyT function, Eq.~(\ref{eq:dyt}).
\end{proof}

\subsection{Theorem 3}
\label{app:theorem3}

\begin{manualtheorem}{3}[RMSNorm Rearrangement]
Let $x \in \mathbb{R}^C$ and 
\begin{align}
y_i &:= \frac{x_i}{\sqrt{\widetilde \sigma^2}}
\tag{\ref{eq:rmsnorm}}
\end{align}
be the output of the RMSNorm with
\begin{align}
    \widetilde \sigma^2 
    &= \frac{1}{C} \sum_{k=1}^C x_k^2
    = \frac{1}{C} \| x \|^2
    \tag{\ref{eq:sigmasq_rmsnorm}}
\end{align}
being the uncentered variance of $x$.
Eq.~(\ref{eq:rmsnorm}) can be written as
\begin{align}
y_i 
 &= \sqrt{C} \cdot \frac{x_i}{\sqrt{\| x_{\backslash i} \|^2 + x_i^2}} \tag{\ref{eq:rmsnorm_alternative}}
\end{align}
where
\begin{align}
    \| x_{\backslash i} \| := \sqrt{ \sum_{k \neq i} x_k^2 }
    \tag{\ref{eq:definition_x_not_i}}
\end{align}
is the 2-norm of the vector $x_{\backslash i} \in \mathbb{R}^{C-1}$ comprised of all channels but $i$.
\end{manualtheorem}
\begin{proof}
\begin{align}
 y_i 
 &\stackrel{(\ref{eq:rmsnorm})}{=} \frac{x_i}{\sqrt{\widetilde \sigma^2}} \nonumber \\
 &\stackrel{(\ref{eq:sigmasq_rmsnorm})}{=} \frac{x_i}{\sqrt{\frac{1}{C} \sum_{k=1}^C x_k^2}} \nonumber \\
 &= \sqrt{C} \cdot \frac{x_i}{\sqrt{\sum_{k \neq i} x_k^2 + x_i^2}} \nonumber \\
 &\stackrel{(\ref{eq:definition_x_not_i})}{=} \sqrt{C} \cdot \frac{x_i}{\sqrt{\| x_{\backslash i} \|^2 + x_i^2}} 
 \tag{\ref{eq:rmsnorm_alternative}}
\end{align}
\end{proof}

\section{Inverse Square Root Unit (ISRU)}
\label{app:isru}

The Inverse Square Root Unit (ISRU) function is defined in \citet{carlile2017improvingdeeplearninginverse} as\footnote{Note that the parameter $\alpha$ is not to be confused with the one used in DyT, Eq.~(\ref{eq:dyt}).}
\begin{align}
    f(x) = \frac{x}{\sqrt{1 + \alpha x^2}}
    \label{eq:isru}
\end{align}
This can also be written as 
\begin{align}
    f(x) 
    &= \frac{x}{\sqrt{\alpha \cdot \left( \frac{1}{\alpha} + x^2 \right)}} \nonumber \\
    &= \frac{\sqrt{\beta} x}{\sqrt{\left( \beta + x^2 \right)}}
    \label{eq:isru_beta}
\end{align}
where we have identified 
\begin{align}
    \beta := \frac{1}{\alpha}
\end{align}
in the second step. Eq.~(\ref{eq:isru_beta}) is the same as DyISRU from Eq.~(\ref{eq:dyisru}) apart from the factor $\sqrt{\beta}$ in the nominator.

\section{DyISRU in Practice}
\label{app:practical_considerations}

\subsection{Initialization of $\beta$}

Fig.~\ref{fig:solutions} illustrates the similarity of DyT and DyISRU if their derivatives at the origin $x_i=0$, given by
\begin{align}
\left. \frac{d y_i}{d x_i} \right|_{x_i = 0} 
&\stackrel{(\ref{eq:dyt})}{=} \alpha 
\label{eq:DyT_slope} \\
\intertext{for DyT and}
\left. \frac{d y_i}{d x_i} \right|_{x_i = 0} 
&\stackrel{(\ref{eq:dyisru})}{=} \frac{1}{\sqrt{\beta}}
\label{eq:DyISRU_slope}
\end{align}
for DyISRU, are the same.
A good starting point for the initialization $\beta_0$ can thus be inferred from values of $\alpha_0$ that have been shown to work well in practice \cite{Zhu2025DyT}, by matching Eq.~(\ref{eq:DyT_slope}) and (\ref{eq:DyISRU_slope}). For example, the commonly used value $\alpha_0 = 0.5$ for DyT corresponds to $\beta_0 = \frac{1}{\alpha_0^2} = 4$ for DyISRU.  

\subsection{Parametrization of $\beta$}

A potential practical problem of the DyISRU function from Eq.~(\ref{eq:dyisru}) is that it is only defined for $\beta > 0$. This can theoretically cause the training to fail, especially if the initial value $\beta_0$ is chosen to be small. A way to avoid instabilities is to parametrize $\beta$ in terms of a differentiable, positive function $f: \mathbb{R} \to \mathbb{R}^+, \widehat \beta \mapsto \beta$, like the softplus function \citep{dugas2000incorporating}:
\begin{align}
\beta &= {\rm softplus}(\widehat \beta) = \log(1+\exp(\widehat \beta))
\label{eq:softplus}
\end{align}

\subsection{Efficiency}

DyT has been shown to provide significant computational speed-ups compared to RMSNorm \cite{Zhu2025DyT}. Since DyISRU is very similar to DyT in that it only uses element-wise transformations, we expect it to be about as fast. However, this question will eventually have to be answered by experiments.

\newpage
\section{Outlier Simulation: Generalization}
\label{app:outlier_simulation_generalization}

We extend the outlier simulation presented in Sec.~\ref{sec:simulations} to other distributions, namely uniform and Laplace, with $\sigma \in \{ 1, 2, 4 \}$. 
Tab.~\ref{tab:outlier_simulation_generalization} shows that the results generalize well.
\begin{table}[!ht]
    \centering
    \scriptsize
    \begin{tabular}{cccccc}
    \toprule
    $\sigma$ & distribution & MAR (DyT) & MAR (DyISRU) & $\alpha$ & $\beta$ \\ \midrule
    1 & normal & 0.276 & 0.002 & 0.096 & 75.1 \\ 
1 & uniform & 0.282 & 0.003 & 0.095 & 77.9 \\ 
1 & Laplace & 0.275 & 0.002 & 0.096 & 74.7 \\ \midrule 
2 & normal & 0.328 & 0.005 & 0.049 & 301.1 \\ 
2 & uniform & 0.327 & 0.007 & 0.048 & 312.3 \\ 
2 & Laplace & 0.327 & 0.005 & 0.049 & 299.4 \\ \midrule 
4 & normal & 0.219 & 0.007 & 0.025 & 1206.6 \\ 
4 & uniform & 0.203 & 0.010 & 0.025 & 1252.6 \\ 
4 & Laplace & 0.221 & 0.007 & 0.026 & 1199.9 \\ \bottomrule
    \end{tabular}
    \caption{Outlier simulation results for different distributions. MAR stands for mean absolute residual, see Sec.~\ref{sec:simulations}.}
    \label{tab:outlier_simulation_generalization}
\end{table}

\end{document}